\documentclass[acmtog,screen,authorversion,nonacm]{acmart}
\AtBeginDocument{%
  \providecommand\BibTeX{{%
    \normalfont B\kern-0.5em{\scshape i\kern-0.25em b}\kern-0.8em\TeX}}}

\usepackage{fancyhdr}
\usepackage{multirow}
\usepackage{tabularx}
\usepackage{array}
\usepackage{algorithm}
\usepackage{algpseudocode}
\usepackage{xcolor}
\usepackage{subfigure}
\usepackage{hyperref}

\begin{document}

\title{Analogist: Out-of-the-box Visual In-Context Learning with Image Diffusion Model}

\author{Zheng Gu}
\email{guzheng@smail.nju.edu.cn}
\orcid{0000-0001-9914-3922}
\affiliation{
  \institution{City University of Hong Kong and State Key Lab for Novel Software Technology, Nanjing University}
  \country{China}
}
\author{Shiyuan Yang}
\email{s.y.yang@my.cityu.edu.hk}
\orcid{0000-0001-8213-5803}
\affiliation{
  \institution{City University of Hong Kong and Tianjin University}
  \country{China}
}
\author{Jing Liao}
\authornote{Jing Liao and Jing Huo are the co-corresponding authors.}
\email{jingliao@cityu.edu.hk}
\orcid{0000-0001-7014-5377}
\affiliation{
  \institution{City University of Hong Kong}
  \country{China}
}
\author{Jing Huo}
\authornotemark[1]
\email{huojing@nju.edu.cn}
\orcid{0000-0002-8504-455X}
\affiliation{
  \institution{State Key Lab for Novel Software Technology, Nanjing University}
  \country{China}
}
\author{Yang Gao}
\email{gaoy@nju.edu.cn}
\orcid{0000-0002-2488-1813}
\affiliation{
  \institution{State Key Lab for Novel Software Technology, Nanjing University}
  \country{China}
}



\begin{abstract}
Visual In-Context Learning (ICL) has emerged as a promising research area due to its capability to accomplish various tasks with limited example pairs through analogical reasoning. However, training-based visual ICL has limitations in its ability to generalize to unseen tasks and requires the collection of a diverse task dataset. On the other hand, existing methods in the inference-based visual ICL category solely rely on textual prompts, which fail to capture fine-grained contextual information from given examples and can be time-consuming when converting from images to text prompts.
To address these challenges, we propose Analogist, a novel inference-based visual ICL approach that exploits both visual and textual prompting techniques using a text-to-image diffusion model pretrained for image inpainting. For visual prompting, we propose a self-attention cloning (SAC) method to guide the fine-grained structural-level analogy between image examples. For textual prompting, we leverage GPT-4V's visual reasoning capability to efficiently generate text prompts and introduce a cross-attention masking (CAM) operation to enhance the accuracy of semantic-level analogy guided by text prompts.
Our method is out-of-the-box and does not require fine-tuning or optimization. It is also generic and flexible, enabling a wide range of visual tasks to be performed in an in-context manner. Extensive experiments demonstrate the superiority of our method over existing approaches, both qualitatively and quantitatively. Our project webpage is available at \href{https://analogist2d.github.io}{https://analogist2d.github.io}.

\end{abstract}




\begin{teaserfigure}
  \includegraphics[width=\textwidth]{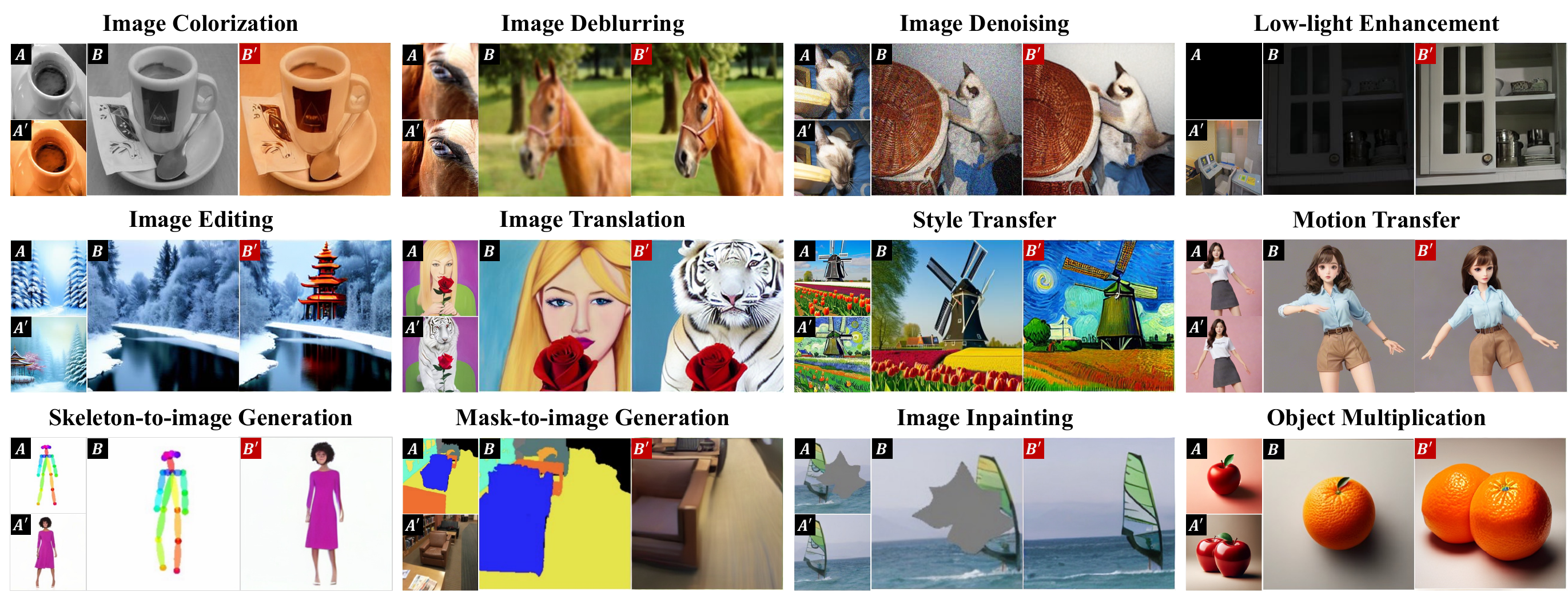}
  \caption{Examples of in-context visual generation by our method using a pretrained Stable Diffusion Inpainting model are demonstrated. With an example image pair $A$ and $A'$, illustrating a visual transformation, and a query image $B$, our method enhances the model's capacity for visual in-context comprehension, producing a reasonable output $B'$ that follows the same visual pattern. Source images: ImageNet~\cite{deng2009imagenet}, LOL~\cite{Chen2018Retinex}, InstructPix2Pix~\cite{brooks2023instructpix2pix}, TongYi QianWen APP, UBC-Fashion~\cite{zablotskaia2019dwnet}, ScanNet~\cite{dai2017scannet}, DAVIS~\cite{perazzi2016benchmark}, DALLE-3~\cite{betker2023improving}.}
  \Description{}
  \label{fig:teaser}
\end{teaserfigure}


\maketitle

\section{Introduction}
As one of the most popular research topics in the recent field of natural language processing (NLP), in-context learning (ICL) represents a paradigm wherein large language models (LLMs) acquire the ability to learn tasks based on a limited set of demonstrative examples~\cite{dong2022survey}. Unlike supervised learning, ICL directly generates predictions using pretrained LLMs~\cite{brown2020language}. This paradigm offers an interpretable interface for interacting with LLMs through language demonstrations, mirroring human decision-making by learning through analogies and similar experiences. ICL significantly lowers computational costs for adapting models to new tasks, making language-model-as-a-service feasible and enabling practical applications in large-scale, real-world tasks such as machine translation~\cite{xu2023small}, information extraction~\cite{He_2023_ICCV}, and complexity reasoning~\cite{wei2022chain}.

Following the success of NLP, research in visual In-Context Learning has entered its embryonic stage of exploration~\cite{yang2023dawn,bai2023sequential}. 
{Specifically, when the demonstration is a pair of images $A$ and $A'$, visual in-context learning can be considered as an image analogy problem~\cite{Hertzmann2001}. This involves analogizing the observed transformation from $A$ to $A'$ and applying it onto a query image $B$, resulting in $B'$. This analogy capability holds significant potential in computer graphics and vision tasks~\cite{vsubrtova2023diffusion,parmar2023zero,cao2023masactrl}.} 
For example, as shown in Figure~\ref{fig:teaser}, with just a single pair of examples without training on a large dataset, the pretrained model can perform tasks ranging from low-level tasks such as colorization, deblurring, denoising, etc., to high-level tasks such as image editing, image translation, motion transfer, etc. Visual ICL also offers significant potential in enhancing creative workflows. Designers can leverage a model to learn design ideas such as color themes, typography, and visual motifs from an example pair and adapt them analogously to different contents.

Existing visual ICL works fall into two categories: training-based and inference-based. Training-based methods train the generative model on diverse in-context tasks~\cite{wang2023incontext,najdenkoska2023context}. The ICL capabilities primarily exhibit tasks similar to their training tasks and have limitations when applied to unseen tasks. Moreover, collecting and organizing the data into in-context task format is laborious. Inference-based methods conduct ICL via appropriate prompting the model during inference, possessing better generalizability. However, existing methods~\cite{vsubrtova2023diffusion,nguyen2023visual} convert the given images into textual prompts, falling short in two aspects. First, the textual prompting is coarse-grained and cannot cover the detailed information presented in the image examples. Second, textual inversion from images requires iterative optimization, which is still time-consuming.

In this work, we propose Analogist, a novel inference-based visual ICL approach, to address the aforementioned challenges. We introduce both visual and textual prompting techniques on a pretrained text-to-image diffusion model. 

Firstly, we introduce a novel visual prompting technique to overcome the coarse-granularity issue in textual prompting. {Inspired by MAEVQGAN~\cite{bar2022visual}, we formulate the ICL task as an image inpainting task by arranging the exemplary image pair $A$ and $A'$, the query image $B$, and the unknown image $B'$ in a $2 \times 2$ grid.} Then, we utilize a pretrained diffusion inpainting model to fill in the region of $B'$. To guide the inpainting process with fine-grained visual contextual information, we propose a self-attention cloning (SAC) method. This method clones the self-attention maps between $A$ and $B$ to the self-attention maps between $A'$ and $B'$ during the forward propagation of the diffusion inpainting model. Since the self-attention maps represent similarity between pixels, the SAC method effectively helps learn structural-level relationships between $A$ and $B$, which are then applied to $A'$ to generate $B'$ analogically.

In addition to visual prompting offering structural-level guidance, we incorporate textual prompting to offer semantic-level guidance by providing appropriate text prompts to the inpainting model. However, unlike previous methods~\cite{vsubrtova2023diffusion,nguyen2023visual} that rely on time-consuming textual inversion optimization, we propose utilizing GPT-4V's visual reasoning capability to analyze the semantic transformation between $A$ and $A'$ and apply it analogically to $B$ to generate a textual description of $B'$. This is facilitated by our well-designed graphical and textual instructions fed into GPT-4V. Furthermore, we introduce a cross-attention masking (CAM) operation to restrict the interaction between text and image to the $B'$ region only, which ensures that the textual prompt more accurately guides the generation of $B'$.

With both semantic-level (coarse-grained) and structural-level (fine-grained) contextual information respectively provided by textual and visual prompting techniques, our approach is capable of performing a wide range of visual tasks in an in-context manner, as illustrated in Figure~\ref{fig:teaser}. Our approach is an out-of-the-box solution that only requires one forward step of a pretrained diffusion model, without the need for fine-tuning or optimization. Extensive experiments and comparisons across different tasks have confirmed that our method outperforms existing training-based and inference-based visual ICL methods, both qualitatively and quantitatively. {Our method is primarily designed for applications where the input $A$ and $A'$ are spatially aligned. Nonetheless, we show that it holds promise for applications in misaligned scenarios as well.} In summary, our contributions can be summarized as follows:

\begin{itemize}
\item We introduce Analogist, an out-of-the-box approach for visual in-context learning that utilizes a pretrained diffusion inpainting model along with effective visual and textual prompting techniques.
\item In visual prompting, we propose a Self-Attention Cloning (SAC) method that effectively guides the image inpainting model to exploit fine-grained contextual information in the $2 \times 2$ grid visual prompt.
\item In textual prompting, we propose to efficiently generate textual prompts using GPT-4V and enhance the accuracy of textual guidance by introducing a Cross-Attention Masking (CAM) operation.
\end{itemize}

\section{Related Work}

\subsection{Visual In-context Learning}

Inspired by the taxonomy in Dong et al. ~\shortcite{dong2022survey}, {we categorize current visual in-context learning into two groups, training-based and inference-based, based on the criterion of whether the model is trained on in-context tasks.}

\paragraph{Training-based Methods} 
Training-based methods train (or finetune) the model on diverse in-context tasks. Painter~\cite{wang2023images} uses paired input and output images as visual prompts to train a Vision Transformer~\cite{dosovitskiy2020image}, which enables the model to learn and perform a wide range of vision tasks. The follow-up work SegGPT~\cite{Wang_2023_ICCV} extends the in-context learning capabilities of Painter specifically for precise and adaptable segmentation across various domains. 
More recently, several work progressively exhibits the ICL ability of state-of-the-art diffusion models~\cite{rombach2022high}. PromptDiffusion~\cite{wang2023incontext} introduces ControlNet~\cite{zhang2023adding} to tune a pretrained Stable Diffusion on six manually designed vision-language tasks. The proposed method is able to generalize to similar, contextually related unseen tasks. However, it poses challenge for users to offer detailed and precise text descriptions.
ImageBrush~\cite{sun2023imagebrush} introduces a novel framework for image manipulation using in-context visual instructions, rather than natural language. An additional prompt encoder is introduced to translate the visual changes depicted in the example images into text features to guide the inpainting model. ImageBrush is built on a diffusion-based inpainting model and trained on several vision datasets.
The above training-based methods necessitate the construction of high-quality and diverse tasks, making the pipeline laborious and inflexible. Meanwhile, the test tasks should ideally bear some similarity to the training tasks, suggesting opportunities for improving generalizability.

\begin{figure*}[t]
    \centering
    \includegraphics[width=\textwidth]{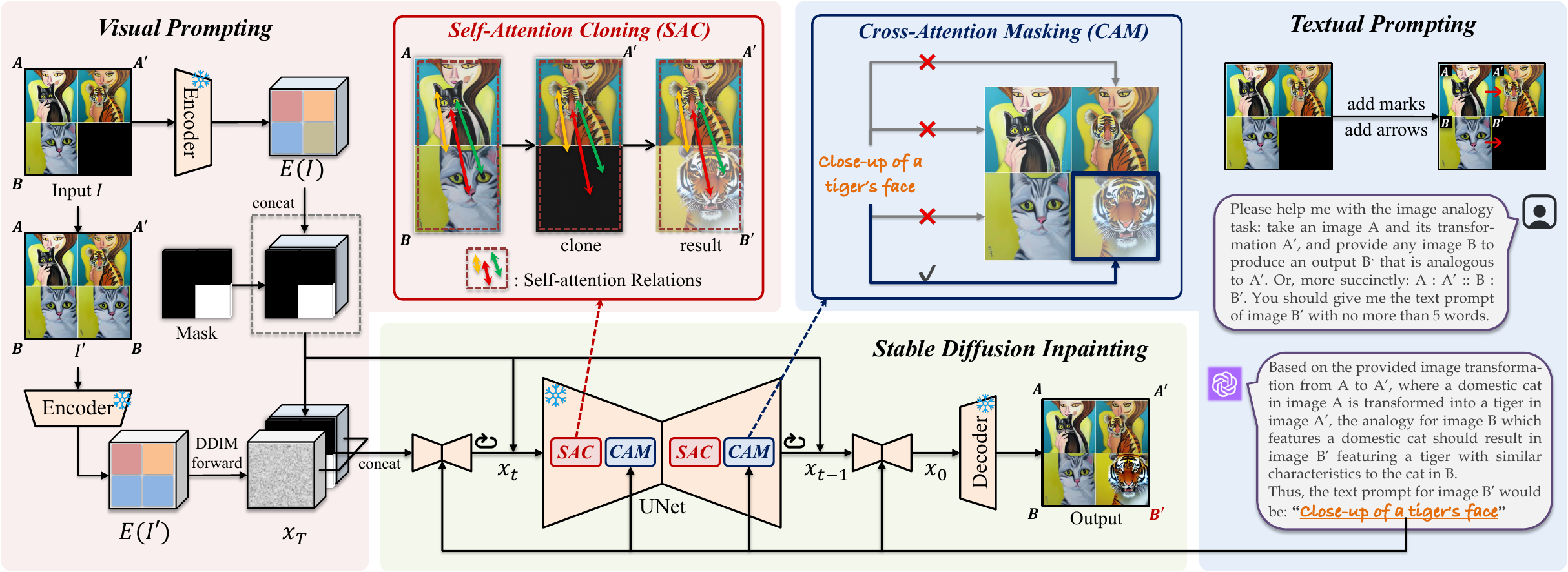}
    \caption{Overview of the proposed Analogist. A visual demonstration is defined by an example pair $A$ (woman holding a cat) and $A'$ (the same woman holding a tiger). Given a new image $B$ (another cat), we format these three images into a $2\times 2$ grid and tackle this problem by fill the missing image via a pretrained Stable Diffusion inpainting model. We employ GPT-4V to provide a proper text description (\textit{i.e.}, ``close-up of a tiger's face'') to further guide the inpainting process. During the process of model inference, Self-Attention Cloning (SAC) and Cross-Attention Masking (CAM) are introduced to encourage the model concentrate on the visual and textual prompts, thus enhance its in-context learning capacities. Source image: InstructPix2Pix~\cite{brooks2023instructpix2pix}.}
    \label{fig:framework}
    \Description{}
\end{figure*}

\paragraph{Inference-based Methods} {Instead of tuning the model parameters, inference-based methods inspire the model's understanding on the given demonstrations during inference time. 
Among them, MAEVQGAN~\cite{bar2022visual} innovatively proposes a visual prompting format of inpainting the missing patch in a $2 \times 2$ grid-like image. The model is pre-trained on figures from computer vision papers which are typically in a regular grid pattern and emerges with ICL capability. However, the generation effects are not entirely satisfactory due to limitations in dataset size and model capacity in comparison with the latest diffusion models.}
VISII~\cite{nguyen2023visual} considers the demonstration as images before and after image editing. This approach estimates the editing instruction based on a pretrained text-based image editing model~\cite{brooks2023instructpix2pix}, producing results with higher quality. However, reverse-engineering the textual description of the differences between two images through optimization remains time-consuming. What's more, by transferring visual information to coarse-grained text, the generation process is merely driven by textual descriptions. The role of visual prompting is not fully leveraged, leading to inaccurate contextual understanding.

Our work falls into the category of inference-based methods and, notably, eliminates the need for additional optimization steps. Instead of solely relying on textual prompts, our approach leverages both textual and visual prompting. This allows us to respectively understand semantic-level and structural-level contextual information for visual ICL. Besides, our method utilizes GPT-4V to get textual prompts instead of textual inversion.

\subsection{Image Analogies}
{Defined by $A:A'::B:B'$, the goal of image analogies~\cite{Hertzmann2001} is to find an ``analogous'' image $B'$ that relates to $B$ in the same way as $A'$ relates to $A$. Such idea can be extended in various ways of image synthesis~\cite{Diamanti_2015,Jamriska_2019,liao2017visual,yuan2024diffmat}.
Recently, DIA~\cite{vsubrtova2023diffusion} investigates the image analogies task with Diffusion model. This method estimates the CLIP features of the given images. The CLIP features are injected into a pretrained text-to-image diffusion model to provide in-context guidance. DIA is capable of executing example-based image editing that encompasses complex, higher-level contextual or structural relationships. However, since the goal of CLIP is to align image and text spaces, the estimated features are high level and struggle to capture detailed image information. }

{Our work aims to tackle the problem of image analogies in the paradigm of visual in-context learning. Different from traditional texture synthesis approaches~\cite{Hertzmann2001,liao2017visual}, the analogy is achieved by prompting a pre-trained text-to-image diffusion model and can be applied to more applications such as low-level tasks, manipulation tasks, and vision tasks.}

\subsection{Prompt-based Image Editing}
{Recent multimodal approaches have demonstrated superior text-image feature alignment capabilities~\cite{radford2021learning,li2022blip}, leading to a series of works on prompt-based image editing. Previous GAN-based methods perform manipulation in the latent space via GAN inversion~\cite{xia2022gan,patashnik2021styleclip,baykal2023clip}. More recent methods utilize text-to-image diffusion models to attain leading outcomes~\cite{cao2023masactrl,brooks2023instructpix2pix,parmar2023zero}. However, these methods struggle to do image analogy task since they take textual descriptions as input, which is not sufficiently intuitive and accurate to depict details related to the image structure. In contrast, our work takes a pair of images as demonstration input, utilizes self-attention to provide structure-related information, and automatically acquires the corresponding textual description through GPT-4V.}

\section{Preliminary}
Since our approach utilizes a pretrained Stable Diffusion inpainting model, we briefly review latent Stable Diffusion in Section~\ref{sec:ldm} as well as the Stable Diffusion inpainting model in Section~\ref{sec:sd_inpainting}.

\subsection{Latent Diffusion Models.}
\label{sec:ldm}
Denoising Diffusion Probabilistic Models (DDPM)~\cite{ho2020denoising} are a class of generative models that gradually convert random noise into structured data through a series of reverse diffusion steps based on a Markov chain. 
Latent Diffusion Models (LDM) like Stable Diffusion (SD)~\cite{rombach2022high} enhances DDPM by employing an encoder $E$ to map high-dimensional data $x$ into lower-dimensional latent space $z=E(x)$. The generation of Stable Diffusion can be guided by an additional text embedding $c(y)$ encoded by CLIP~\cite{radford2021learning} and a text prompt $y$. During training, an UNet model, parameterized by $\theta$, is optimized to eliminate the noise $\epsilon$ introduced into $z_t$:
\begin{equation}
    \mathcal{L}=\mathbb{E}_{z \sim E(x),y,\epsilon \sim \mathcal{N}(0,1),t} \left [ {\left \| { \epsilon-\epsilon_{\theta}(z_t,t,c(y)) } \right \|}^2_2 \right ].
\end{equation}
During inference, a randomly sampled latent $z_T \sim \mathcal{N}(0,1)$ is progressively denoised through the model to produce a clean latent representation $z_0$ by 
\begin{equation}
    z_{t-1} = \frac{1}{\sqrt{\alpha_t}}\left [ z_t - \frac{1-\alpha_t}{1-\sqrt{\bar{\alpha}_t}} \epsilon_{\theta}\left ( z_t, t, c(y) \right ) \right ],
\end{equation}
where $\bar{\alpha}_t = \prod_{i=1}^{t} \alpha_t$. Subsequently, the clean latent is fed into the decoder to obtain the generated image $D(z_0)$.


\subsection{Stable Diffusion Inpainting Model}
\label{sec:sd_inpainting}
We apply our method over the pretrained Stable Diffusion inpainting model, which is fine-tuned to boasts an additional feature of image inpainting. The forward process of the inpainting pipeline is as follows:
\begin{equation}
    z_{t-1} = \frac{1}{\sqrt{\alpha_t}}\left [ z_t - \frac{1-\alpha_t}{1-\sqrt{\bar{\alpha}_t}} \epsilon_{\theta}\left ( z_t, t, c(y), E(I_m), M \right ) \right ],
\end{equation}
The UNet is updated to include five extra input channels – four dedicated to the encoded masked-image $E(I_m)$ and one for the mask $M$ itself. These two extra inputs are concated with $z_t$ to fed into the UNet to predict the noise at each time step.

\section{Method}
The goal of ICL is to encourage pretrained model to learn tasks given only a few examples in the form of demonstration~\cite{dong2022survey}. Specific to the image domain, the demonstration is defined as an example image pair $A$ and $A'$, where $A'$ is the result obtained by applying a certain visual effect or transformation to $A$. Given a new query image $B$, the model is expected to apply the same effect to $B$, thus creating a new image $B'$, so that $A:A'::B:B'$~\cite{Hertzmann2001}. This process demonstrates the model's understanding and replication of visual transformations from a given demonstration to a new context, exhibiting the ICL ability. 

As illustrated in Figure~\ref{fig:framework}, to address this issue, we approach it from both visual structural-level (Section~\ref{sec:visual_prompt}) and textual semantic-level (Section~\ref{sec:text_prompt}) perspectives. For visual prompting (red region in Figure~\ref{fig:framework}), we formulate the input images into a 2x2 grid image, utilizing a pretrained diffusion inpainting model to fill in the missing region in Section~\ref{sec:2x2grid}. To introduce more fine-grained visual information, we propose Self-Attention Cloning (SAC) in Section~\ref{sec:sac}. For textual prompting (blue region in Figure~\ref{fig:framework}), GPT-4V is elaborated to provide semantic-level guidance to the generation process in Section~\ref{sec:gpt4v}. To foster semantic correspondence between the inpainted image and the text prompt, we propose Cross-Attention Masking (CAM) in Section~\ref{sec:cam}.



\begin{figure}[t]
    \centering
    \includegraphics[width=0.99\linewidth]{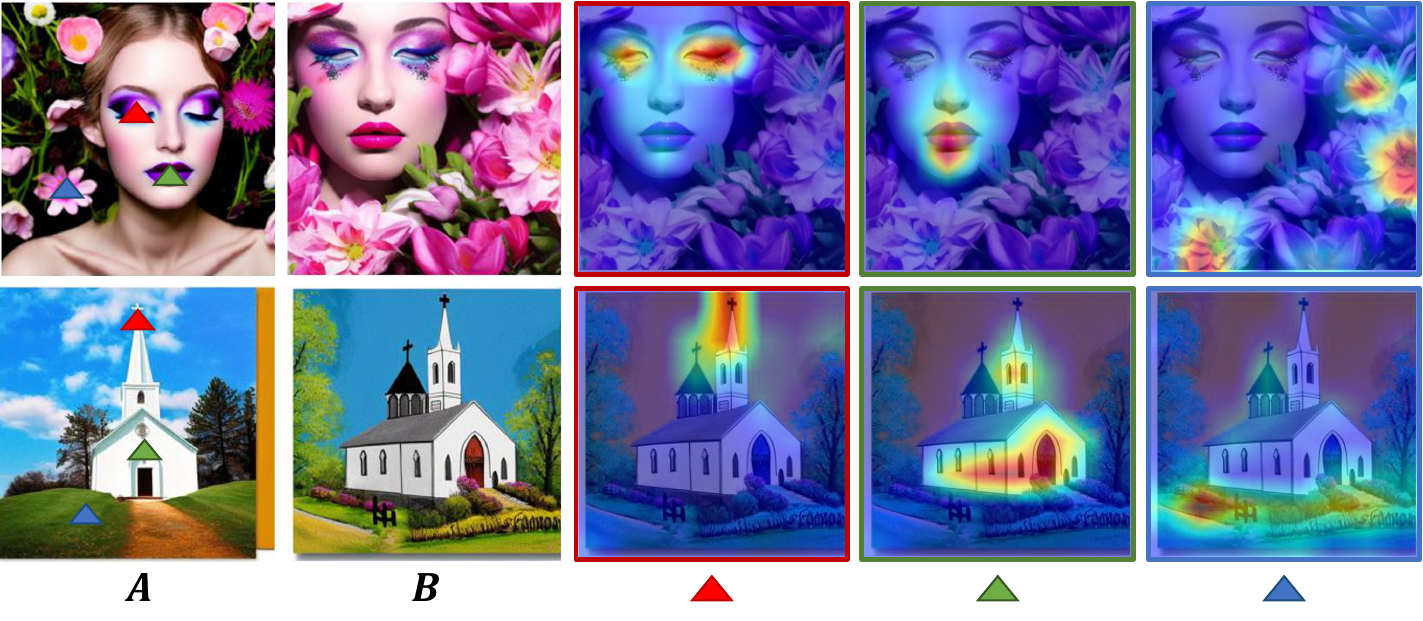}
    \caption{Visualization of the attention relationships. Given an anchor point on image $A$ (shown in red, green, and {blue} colors), we calculate the attention values between this point and all regions of image $B$. Soucre image: InstructPix2Pix~\cite{brooks2023instructpix2pix}.}
    \Description{}
    \label{fig:attention}
\end{figure}

\subsection{Visual Prompting}
\label{sec:visual_prompt}
To introduce fine-grained structural-level visual guidance in the in-context inference process, we construct a visual prompt in the form of a $2\times 2$ grid-like image for the pretrained inpainting model, and provide visual contextual information by cloning the self-attention associations between the given images.

\subsubsection{$2\times 2$-grid Prompting}
\label{sec:2x2grid}
Image inpainting models fill in unknown areas of an image based on its known regions, which naturally aligns with the concept of ICL. As shown in Figure~\ref{fig:framework}, to take advantage of this property, we first rearrange the input images $A$, $A'$, and $B$ into a single $2\times2$ grid-like image, denoted as $I$. Image $B$ is pasted to the bottom right corner of the grid image, getting image $I'$. We extract the features of the pasted image, $E(I')$, and add noise to it via diffusion forward process, getting the initial $x_T$. To align with the interface of the pretrained model, a mask image $M$ is simultaneously generated. In this mask, the bottom right region is entirely ones, while the remaining regions are zeros. {At each timestep $t$, the latent $x_t \in \mathbb{R}^{b \times 4 \times h \times w}$ is concatenated with the feature $E(I) \in \mathbb{R}^{b \times 4 \times h \times w}$ and mask $M \in \mathbb{R}^{b \times 1 \times h \times w}$, constructing the input of the UNet. By establishing such a $2\times 2$-grid prompt, we encourage the model to fill in the content of unknown area ($B'$) based on the contextual regions ($A$, $A'$, and $B$) in the image.}

\begin{figure}[t]
    \centering
    \includegraphics[width=0.99\linewidth]{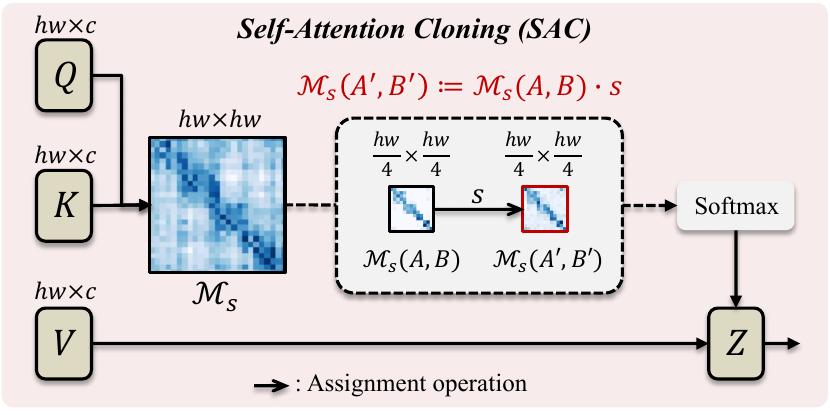}
    \caption{Detailed illustration of self-attention cloning (SAC). The sub self-attention map $\mathcal{M}_s(A',B')$ is set as the value of $\mathcal{M}_s(A,B)$, denoting cloning the relation between $A$ and $B$ to that of $A'$ and $B'$.}
    \Description{}
    \label{fig:sac}
\end{figure}

\subsubsection{Self-Attention Cloning}
\label{sec:sac}
The key of in-context learning is to recognize task instruction from the given demonstration. Previous inference-based work extract the visual instructions through cross-attention injection, which could only provides coarse and imprecise guidance. Differently, we introduce fine-grained structural-aware contextual information via self-attention.

Our motivation comes from an observation that the Diffusion model accurately constructs associations between different positions in the known areas through self-attention. We show the visualization of self-attention relations in Figure~\ref{fig:attention}. We calculate the attention values between key semantic positions (\textit{e.g.}, the eyes, mouth, and flower in the first row and the spire, building, and the background grassland in the second row) in $A$ and all regions in $B$. The results demonstrate that the visual associations between images can be accurately identified through self-attention, which could be more accurate than abstract semantic text prompts as guidance.
{Based on this observation, we propose to use self-attention as a structural-level prior to guide the in-context generation procedure by modulating self-attention in UNet. We show an example in Figure~\ref{fig:framework} of translating a cat into a tiger. The relative positional relationship of the tiger in $B'$ and the tiger in $A'$ should be consistent with the relative positional relationship of the two cats in $B$ and $A$.}

We present detailed illustration of the proposed self-attention cloning (SAC) in Figure~\ref{fig:sac}. Denote the image feature before self-attention as $F_i \in \mathbb{R}^{h \times w \times c}$. The self-attention map $\mathcal{M}_s \in \mathbb{R}^{hw \times hw}$ records the similarity of each position on the entire image with other positions, which also includes the similarities between $A$ and $B$, as well as between $A'$ and $B'$. We extract the sub self-attention map $\mathcal{M}_s(A,B) \in \mathbb{R}^{\frac{hw}{4} \times \frac{hw}{4}}$ and assign its value to $\mathcal{M}_s(A',B') \in \mathbb{R}^{\frac{hw}{4} \times \frac{hw}{4}}$:
\begin{equation}
    \mathcal{M}_s(A',B') := \mathcal{M}_s(A,B) \cdot s,
\end{equation}
where $s$ is a coefficient used to balance the degree of preserving the structure of image $B$ and the degree of applying transformations. We perform the self-attention cloning operation before softmax to prevent the original self-attention results being excessively affected.

\begin{figure}[t]
    \centering
    \includegraphics[width=0.99\linewidth]{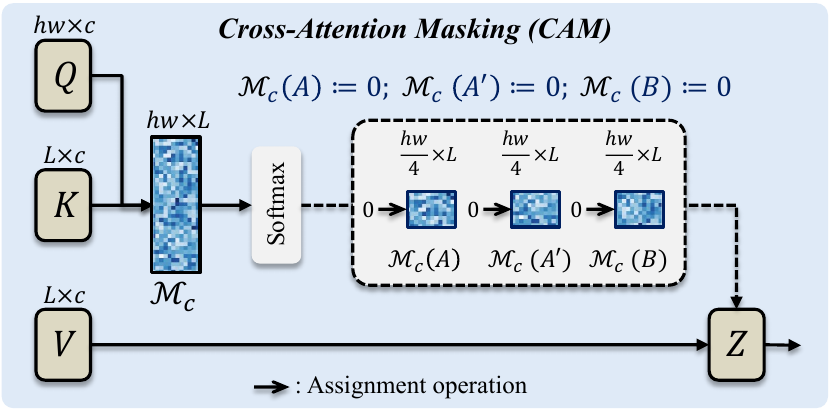}
    \caption{Detailed illustration of cross-attention masking (CAM). The sub cross-attention map between text embedding and regions $A$, $A'$, and $B$ are set to zero, making the semantic guidance more focused on region $B'$.}
    \Description{}
    \label{fig:cam}
\end{figure}

\subsection{Textual Prompting}
\label{sec:text_prompt}
Cloning self-attention effectively manages basic in-context visual guidance, yet the diffusion model's celebrated text-to-image feature remains underutilized to provide semantic-level guidance. To address this, we utilize GPT-4V's visual reasoning abilities~\cite{yang2023dawn} to provide semantic guidance to the inpainting model. 

\subsubsection{GPT-4V Prompting}
\label{sec:gpt4v}
We prompt GPT-4V to generate a coherent text description to aid the inpainting process. Considering the consistency of the entire pipeline, we feed the whole $2x2$ grid-like image directly into GPT-4V with a pre-designed problem Description, as depicted in Figure~\ref{fig:framework}. We employ two carefully-designed graphical instructions to make it easier for GPT-4V to understand the task. Firstly, inspired by~\cite{yang2023set}, we place a letter mark ($A$, $A'$, $B$, $B'$) in the top-left corner of each grid cell. Secondly, we add prominent arrow markers ($\rightarrow$) between $A$ and $A'$, as well as between $B$ and $B'$, to indicate the relationship between the two images. These approaches introduce structured, easily identifiable reference points, facilitating more effective and accurate responses to queries involving visual content. Then, GPT-4V is asked to perform an analogy and output the text description for $B'$. Finally, we use GPT-4V's answer as the semantic-level positive text prompt to reinforce the model's ICL capabilities.
We also employ negative text prompts (\textit{i.e.}, ``Messy, Disordered, Chaotic, Cluttered, Haphazard, Unkempt, Scattered, Disheveled, Tangled,  Random'') to prevent the diffusion model from generating irregular and illogical results. These two prompts work cooperatively to inject semantic-level guidance into the model.

\begin{figure*}[t]
    \centering
    \includegraphics[width=0.94\textwidth]{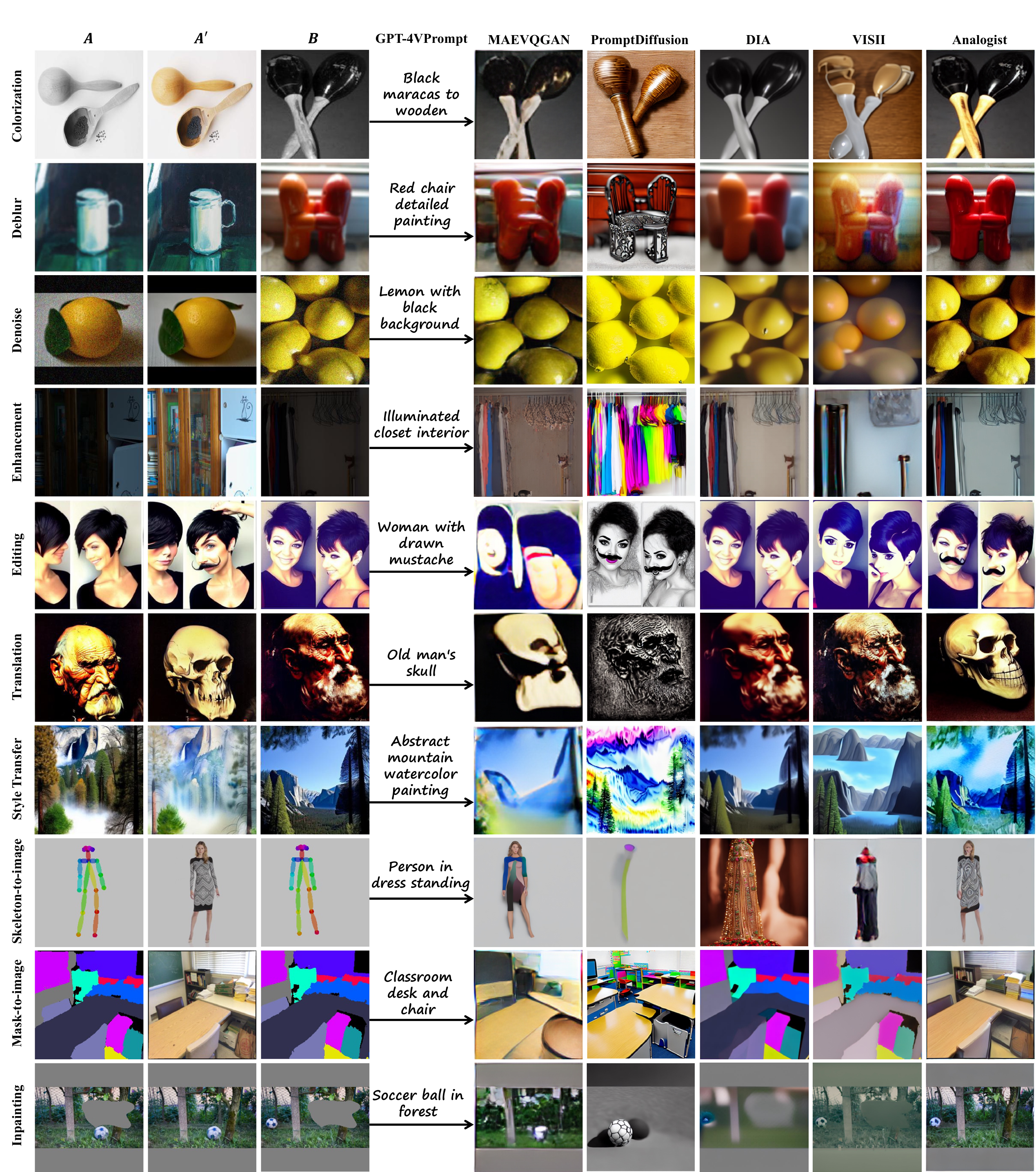}
    \caption{Comparison with other baseline methods, each row indicates one task, given the input image pair $A$, $A'$ and query image $B$. Since MAEVQGAN~\cite{bar2022visual} does not take text as input and DIA~\cite{vsubrtova2023diffusion} and VISII~\cite{nguyen2023visual} estimate the text prompts by extra optimization, the text prompts generated by GPT-4V prompting are only used by PromptDiffusion~\cite{wang2023incontext} and Analogist. Source images: ImageNet~\cite{deng2009imagenet}, LOL~\cite{Chen2018Retinex}, InstructPix2Pix~\cite{brooks2023instructpix2pix}, UBC-Fashion~\cite{zablotskaia2019dwnet}, ScanNet~\cite{dai2017scannet}, DAVIS~\cite{perazzi2016benchmark}.}
    \Description{}
    \label{fig:comparison}
\end{figure*}

\begin{figure*}[t]
    \centering
    \includegraphics[width=0.98\textwidth]{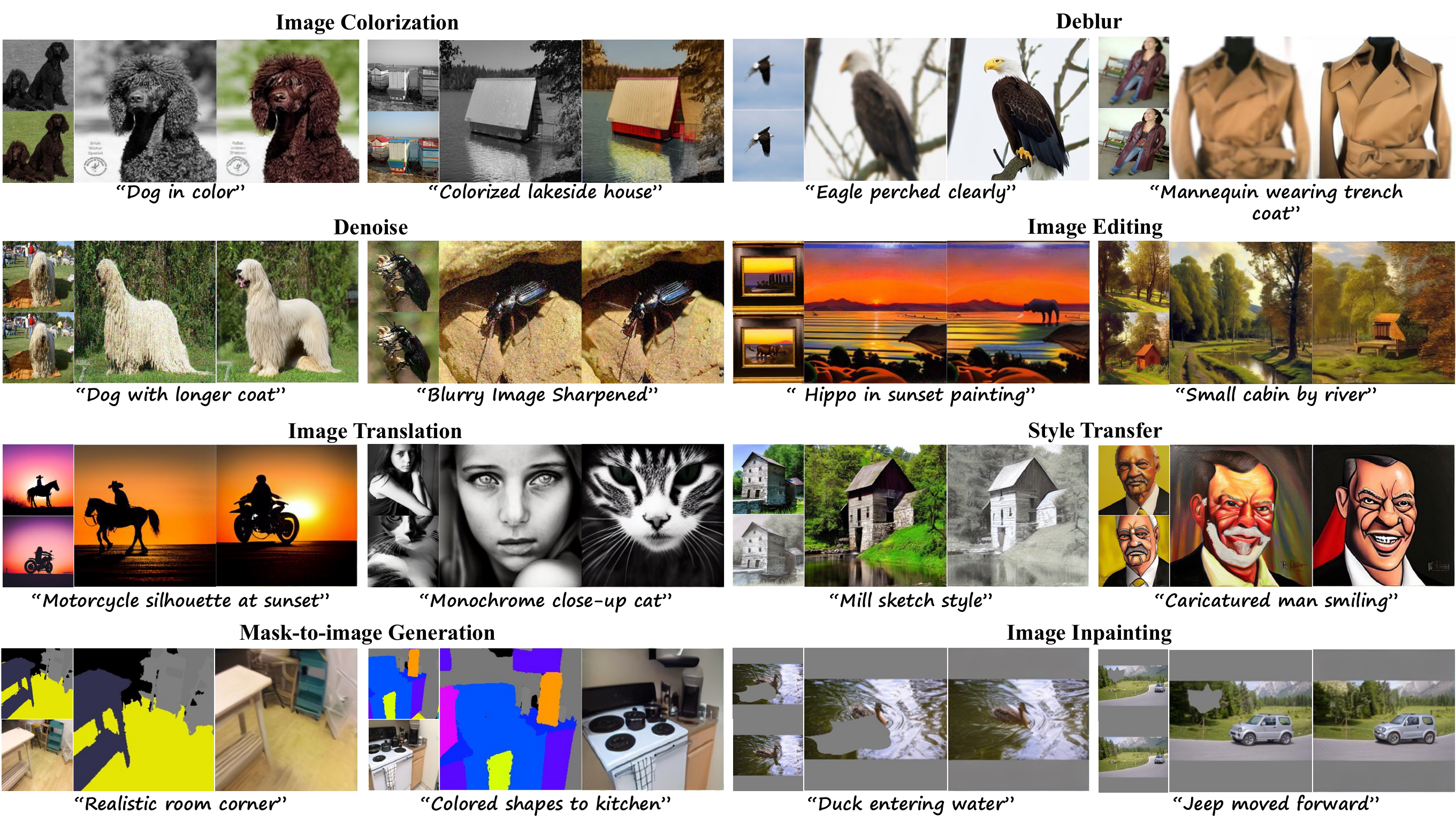}
    \caption{Examples of results generated by the proposed Analogist on different tasks. In each example, the image $A$ and $A'$ are shown in the first column, the image $B$ and generated image $B'$ is shown in the second and third column. The text prompt generated via GPT-4V is shown below each example. Source ImageNet~\cite{deng2009imagenet}, InstructPix2Pix~\cite{brooks2023instructpix2pix}, ScanNet~\cite{dai2017scannet}, DAVIS~\cite{perazzi2016benchmark}.}
    \Description{}
    \label{fig:more_results}
\end{figure*}

\begin{figure}[t]
    \centering
    \includegraphics[width=\linewidth]{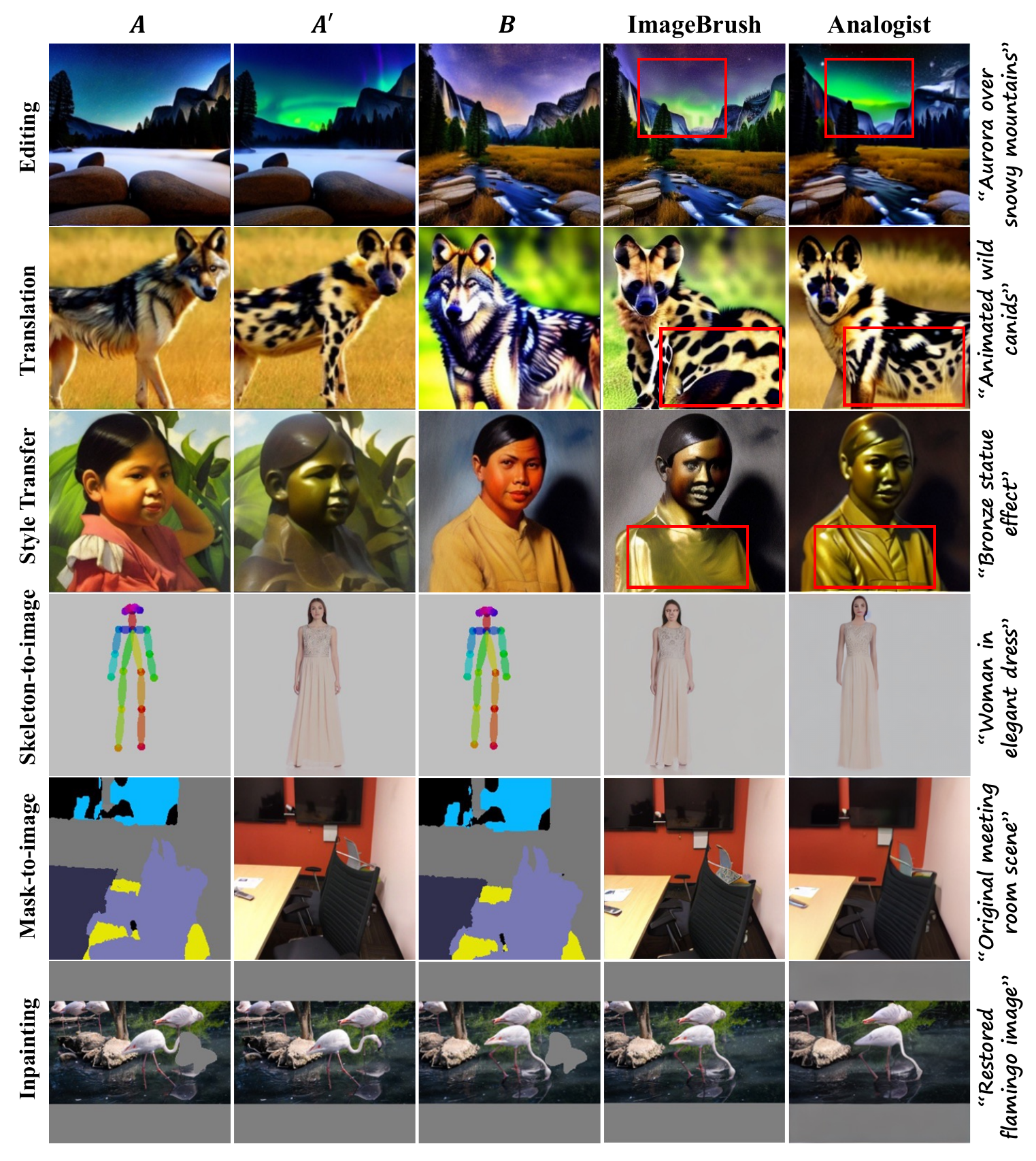}
    \caption{Comparison with ImageBrush~\cite{sun2023imagebrush}. The result of ImageBrush in the first three tasks are from the original paper and the result of the last three tasks are provided by the authors of ImageBrush. Source images: InstructPix2Pix~\cite{brooks2023instructpix2pix}, UBC-Fashion~\cite{zablotskaia2019dwnet}, ScanNet~\cite{dai2017scannet}, DAVIS~\cite{perazzi2016benchmark}.}
    \Description{}
    \label{fig:compare_ib}
\end{figure}

\subsubsection{Cross-Attention Masking} 
\label{sec:cam}
Note that the prompt obtained from GPT-4V is specifically tailored for $B'$, yet the textual guidance impacts the entire image through cross-attention in the UNet. To address this issue, we propose cross-attention masking (CAM): in cross-attention layers, we restrict the text interacts only with the region corresponding to $B'$. Specifically, suppose the cross-attention map as $\mathcal{M}_c \in \mathbb{R}^{hw \times L}$, where $L$ denotes the length of text embedding. We repurpose the indices of different regions identified in the previous SAC process and set the attention values between the text and regions other than $B'$ (\textit{i.e.}, $A$, $A'$, and $B$) to zero:
\begin{equation}
    \mathcal{M}_c(A):=0; \mathcal{M}_c(A'):=0; \mathcal{M}_c(B):=0.
\end{equation}
As illustrated in Figure~\ref{fig:cam}, we utilize the attention map post-softmax, as we are completely obstructing the relationship between the text and regions outside of $B'$.

As for the attention map indexing in SAC and CAM, due to the fixed positions of each image, we are able to pre-calculate the indices required for extracting the necessary sub-attention maps (\textit{e.g.}, $\mathcal{M}_s(A,B)$ and $\mathcal{M}_c(A)$) from the entire attention map. This pre-determination streamlines the entire pipeline, enhancing its simplicity and efficiency. 

\section{Experiments}
\subsection{Implementation Details}
{We implement our work in PyTorch~\cite{paszke2019pytorch}. The input images $A$, $A'$, $B$ are resized to $256 \times 256$ and spatially combined to form a $512\times 512$ grid-like image. We used a publicly available Stable Diffusion inpainting model\footnote{https://huggingface.co/runwayml/stable-diffusion-inpainting}. The model is initialized with SD1.2 and trained on inpainting task, therefore capable of inpainting the missing areas specified by a mask. The UNet architecture contains 16 blocks, each consists of one cross-attention and one self-attention. We perform SAC and CAM from layer 3 to 10 at all timesteps in the diffusion process. The scale for classifier-free guidance is set at $15$. The coefficient for self-attention cloning $s=1.3$ in all experiments except for skeleton-to-image where $s=1.4$. All experiments are conducted on an RTX 3090 GPU.}

\subsection{Evaluation Setup}
\paragraph{Dataset} We employ the following three major categories, totaling ten tasks to evaluate the effectiveness of the proposed method quantitatively: low-level tasks, manipulation tasks, and more challenging vision tasks.
\begin{itemize}
    \item \textbf{Low-level tasks.} We test out method on four low-level tasks, \textit{i.e.}, image colorization, image deblurring, image denoising, and image enhancement. For the first three tasks, we sample in-the-wild images from ImageNet~\cite{deng2009imagenet} and apply corresponding transformations (\textit{i.e.}, grayscale, gaussian blurry, adding noise). For image enhancement, we use the LOL dataset~\cite{Chen2018Retinex}, which consists of low/normal-light image pairs. We collect 100 samples for each low-level task.
    \item \textbf{Manipulation tasks.} We select three kind of image manipulation tasks (\textit{i.e.}, image editing, image translation, and style transfer) from the CLIP-filtered subset processed by InstructPix2Pix~\cite{brooks2023instructpix2pix}. Since the dataset is constructed for general image editing, we split the samples into three tasks based on the keywords. Instructions containing ``add'', ``remove'' are considered as image editing tasks, those with ``make, turn, change'' are image translation tasks. Each manipulation task contains 200 samples.
    \item \textbf{Vision tasks.} We select three more challenging vision tasks for evaluation: skeleton-to-image generation from UBC-Fas-hion~\cite{zablotskaia2019dwnet}, mask-to-image generation from ScanNet~\cite{dai2017scannet}, and image inpainting from DAVIS dataset~\cite{perazzi2016benchmark}. Each task contains 200 samples.
\end{itemize}
By developing these three major categories, we can evaluate if the pretrained model is capable of understanding, processing, and utilizing visual information across various levels, while also evaluating its ability to generalize effectively across these tasks.

\paragraph{Baseline methods} We take four methods, MAEVQGAN~\cite{bar2022visual}, PromptDiffusion~\cite{wang2023incontext}, DIA~\cite{vsubrtova2023diffusion} and VISII~\cite{nguyen2023visual} as our baseline. All baseline methods utilize the official implementations and checkpoints provided. Since PromptDiffusion~\cite{wang2023incontext} requires text as part of its input, but most of the test datasets (such as low-level) do not have paired text descriptions, we input the same text prompts as ours that obtained from GPT-4V into PromptDiffusion~\cite{wang2023incontext} to ensure a fair comparison.

\paragraph{Evaluation Metrics} We evaluate the model's ICL capacity via the CLIP direction similarity between the demonstration and the produced results. We utilize the Image Encoder from CLIP to extract the image features of $A$, $A'$, $B$, and the generated $B'$. Then, we calculate the cosine similarity between the directional changes from $A$ to $A'$ and from $B$ to $B'$. The higher the similarity, the more consistent the inferred $B'$ is with the transformation effects applied to $A$. Due to the generation diversity of diffusion models, we do not compare pixel-level metrics like SSIM and PSNR. Instead, we calculate FID between the generated $B'$ images and the ground truth images. In order to obtain more accurate result, we merge all the data in each major category to calculate the FID values for comparison.

\subsection{Qualitative Results}
Figure~\ref{fig:comparison} presents comparison of our method with the baselines on all of the ten tasks. For MAEVQGAN~\cite{bar2022visual}, due to the lack of specific structuring of training data into the form of tasks and the absence of textual guidance, the quality of the generated output is relatively poor, especially for high-level tasks like manipulation. 
For PromptDiffusion~\cite{wang2023incontext}, the bias in training task (\textit{i.e.}, image-to-HED, HED-to-image) significantly impacts the ICL generalizability of the model. As shown in the example of deblur and translation, the results tend to produce line drawings similar with edge detection results.
For the other two inference-based methods DIA~\cite{vsubrtova2023diffusion} and VISII~\cite{nguyen2023visual}, they conduct in-context learning through the estimated text solely, making it difficult provide sufficiently accurate prompt information to generate the correct results. Our method takes into account guidance at both the visual and semantic levels, which can produce accurate and reasonable in-context outputs. Notice that GPT-4V prompting may struggle with vision tasks, giving coarse descriptions. For example, ``person in dress standing'' in the skeleton-to-image example does not give the detailed description that what pose the woman should be standing in. However, thanks to the proposed SAC operation, these structure-aware in-context information can be still captured and utilized to produce the correct results. Figure~\ref{fig:more_results} shows further results of Analogist on these tasks, demonstrating the ICL capabilities of our proposed method. More randomly selected results are shown in supplementary materials.

Additionally, we conducted a comparison with ImageBrush~\cite{sun2023imagebrush}. Since ImageBrush has not {released} the code, the comparison is made in the range of training tasks of ImageBrush. As shown in Figure~\ref{fig:compare_ib}, it is worth noting that our method is more effective at preserving the details in Image $B$. Especially in manipulation tasks, the color of the aurora, the contour structure of the animals, and the texture on the clothing are better preserved. This is because our proposed visual and textual prompting contain more detailed in-context information. On the three vision tasks, we achieve competitive results with ImageBrush. Note that our model is not fine-tuned specifically for these tasks, which demonstrate our superiority of in-context generalizability as an inference-based method.

\begin{table*}[t]
\centering
\caption{Quantitative comparison on different category of tasks with previous ICL approaches. We report the cosine similarity between CLIP direction from $A$ to $A'$ and from $B$ to $B'$. Higher similarity represents more contextually appropriate generated results. The best results are highlighted.}
\begin{tabularx}{\textwidth}{ll>{\centering\arraybackslash}X>{\centering\arraybackslash}X>{\centering\arraybackslash}X>{\centering\arraybackslash}X>{\centering\arraybackslash}X}
\toprule
Category                            & Task              & MAEVQGAN              & PromptDiffusion       & DIA     & VISII               & Analogist          \\ \hline
\multirow{4}{*}{Low level tasks}    & Colorization      & 0.0558                & {0.1283}    & 0.0066  & 0.1061              & \textbf{0.1797}   \\
                                    & Deblur            & -0.0961               & {0.0251}    & -0.1337 & 0.0081              & \textbf{0.0608}   \\
                                    & Denoise           & -0.0389               & {0.1612}    & 0.1212  & 0.1098              & \textbf{0.2391}   \\
                                    & Enhancement       & 0.1120                & 0.1551                & -0.1443 & {0.2181}  & \textbf{0.2251}   \\ \hline
\multirow{3}{*}{Manipulation tasks} & Image Editing     & 0.1600                & 0.1768                & 0.0922  & \textbf{0.2181}     & {0.1800}\\
                                    & Image Translation & 0.2526                & 0.2426                & 0.1617  & {0.2965}  & \textbf{0.3136}   \\
                                    & Style Transfer    & 0.2274                & 0.2336                & 0.1515  & \textbf{0.2687}     & {0.2455}\\ \hline
\multirow{3}{*}{Vision tasks}       & Skeleton-to-image & 0.4452                & {0.6150}    & 0.2874  & 0.5201              & \textbf{0.7334}   \\
                                    & Mask-to-image     & {0.4467}    & 0.3984                & 0.1590  & 0.3071              & \textbf{0.5531}   \\
                                    & Inpainting        & -0.0357               & 0.0014                & -0.0511 & {0.0619}  & \textbf{0.1013}   \\ \hline
Average                             &                   & 0.1529                & 0.2137                & 0.0650  & 0.2104              & \textbf{0.2832}   \\ \bottomrule
\end{tabularx}
\label{tab:clip_dir}
\end{table*}

\begin{table}[t]
\centering
\caption{Comparison of FID between the generated $B'$s and the ground-truth images. The best results are highlighted. Our method outperforms previous methods in terms of all the three task categories.}
\begin{tabular}{lccc}
\toprule
{Method}        & {Low-level}           & {Manipulation}    & {Vision}              \\
\midrule
MAEVQGAN        & 181.48                & 143.19            & 169.74                \\
PromptDiffusion & 180.39                & 111.79            & 159.02                \\
DIA             & 173.10                & 103.39            & 191.51                \\
VISII           & {140.39}    & {88.36} & {138.44}    \\
Analogist        & \textbf{114.15}       & \textbf{85.67}    & \textbf{96.67}        \\
\bottomrule
\end{tabular}
\label{tab:fid}
\end{table}

\begin{table}[t]
\centering
\caption{User study results. In each task, we report the average percentage of selected result by the users. The best results are highlighted. Our approach garnered the highest number of selections.}
\begin{tabular}{lccc}
\toprule
{Method}        & {Low-level}           & {Manipulation}    & {Vision}              \\
\midrule
MAEVQGAN        & 3.51\%                & 3.45\%            & 0.87\%                \\
PromptDiffusion & 5.33\%                & 14.99\%            & 9.09\%                \\
DIA             & 4.88\%                & 3.32\%            & 0.43\%                \\
VISII           & {20.18\%}             & {18.30\%}          & {15.58\%}    \\
Analogist        & \textbf{66.10\%}       & \textbf{59.95\%}    & \textbf{74.03\%}        \\
\bottomrule
\end{tabular}
\label{tab:user_study}
\end{table}

\subsection{Quantitative Comparisons}
\paragraph{CLIP Direction} We compute the following CLIP direction similarity, $cos[(\mathcal{E}(B')-\mathcal{E}(B)), (\mathcal{E}(A')-\mathcal{E}(A))]$, to evaluate how faithfully the transformations provided by the model adhere to the transformations contained in the given examples. The results are shown in in Table~\ref{tab:clip_dir}. Note that VISII~\cite{nguyen2023visual} achieves acceptable results in manipulation tasks since the model it utilizes is pretrained on this ip2p dataset~\cite{brooks2023instructpix2pix}. Overall, our method demonstrates superior ICL capabilities across all these tasks.

\paragraph{Fréchet inception distance (FID)} We calculate FID between generated images and ground truth on the entire major category. The results are shown in Table~\ref{tab:fid}. The proposed Analogist outperforms all baselines across the three major tasks. {Notice that VISII~\cite{nguyen2023visual} outperforms other baselines on manipulation tasks. This is because VISII leverages an InstructPix2Pix~\cite{brooks2023instructpix2pix} model which is pretrained on the same dataset, making it more familiar with generating data of similar quality.}

\paragraph{User Study} We conduct a user study to evaluate the perceptual performance of our method. The user study consisted of 50 questions, with 42 participants involved, containing all of the 10 kind of tasks. In each question, first, we presented the participants with images $A$ and $A'$, asking them to analyze the changes between them. Then, we provided image $B$ and tasked them with predicting the expected transformation of $B$ following the same pattern. Subsequently, we displayed the outputs generated by different methods for this task, and the participants were required to select the one they deemed most consistent with the identified pattern and of the highest generative quality. We report the average selection result for the three major tasks: low-level tasks, manipulation tasks, and vision tasks in Table~\ref{tab:user_study}. Our proposed method exhibited the highest rate of being chosen among all of the three tasks.

\begin{figure*}[t]
    \centering
    \includegraphics[width=\textwidth]{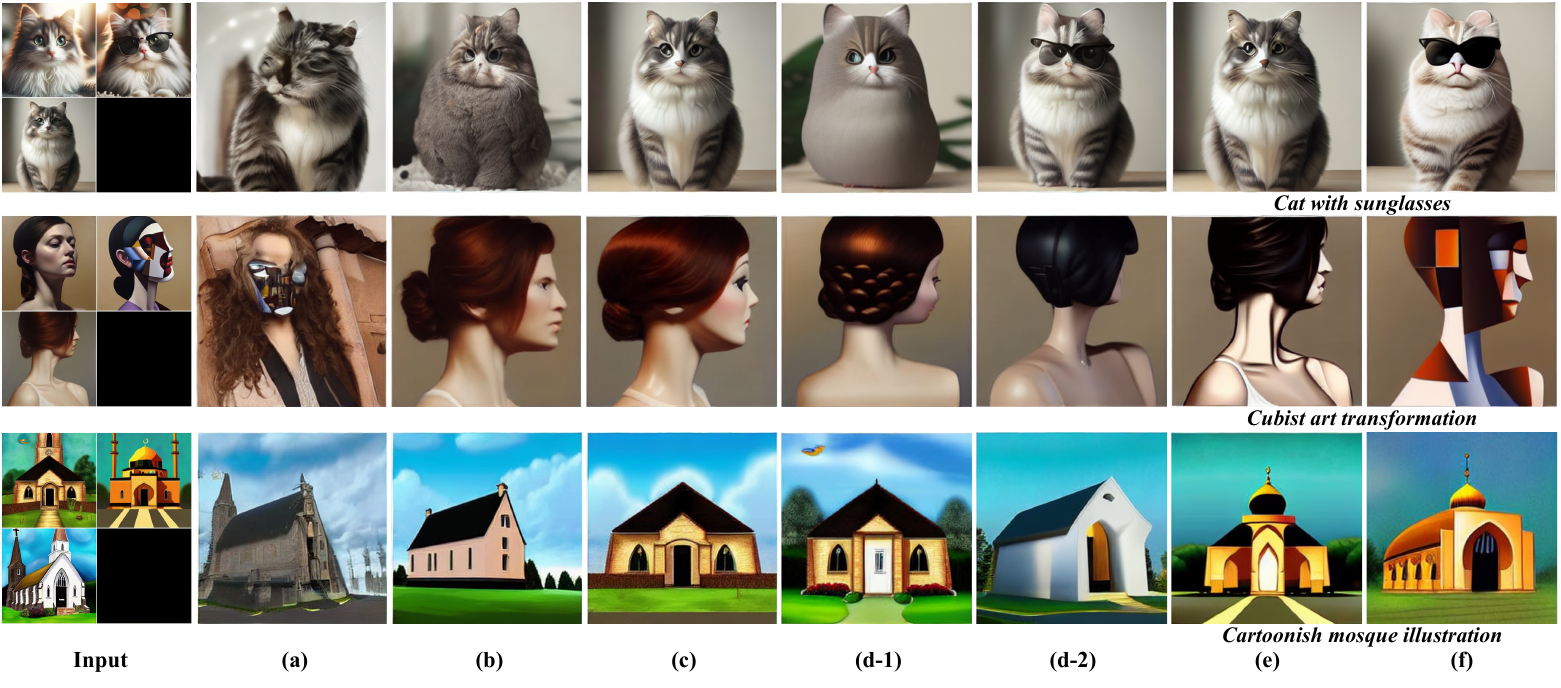}
    \caption{Ablation on the proposed components. An input $2\times2$ image grid is inpainted by: (a) pretrained SD Inpainting model with random noise as input, (b) initializing $B'$ as noised $B$, (c) adding negative prompt, (d-1) adding self-attention cloning (SAC) by $\mathcal{M}_s(B,B'):=\mathcal{M}_s(A,A')$, (d-2) adding SAC by $\mathcal{M}_s(A',B'):=\mathcal{M}_s(A,B)$, (e) adding GPT-4V prompting without cross-attention masking (CAM), and (f) adding CAM (the full approach). Source images: The $1^{st}$ row are generated by DALLE-3~\cite{betker2023improving} and all others are from InstructPix2Pix~\cite{brooks2023instructpix2pix}.}
    \Description{}
    \label{fig:ablation}
\end{figure*}

\begin{figure}[t]
    \centering
    \includegraphics[width=\linewidth]{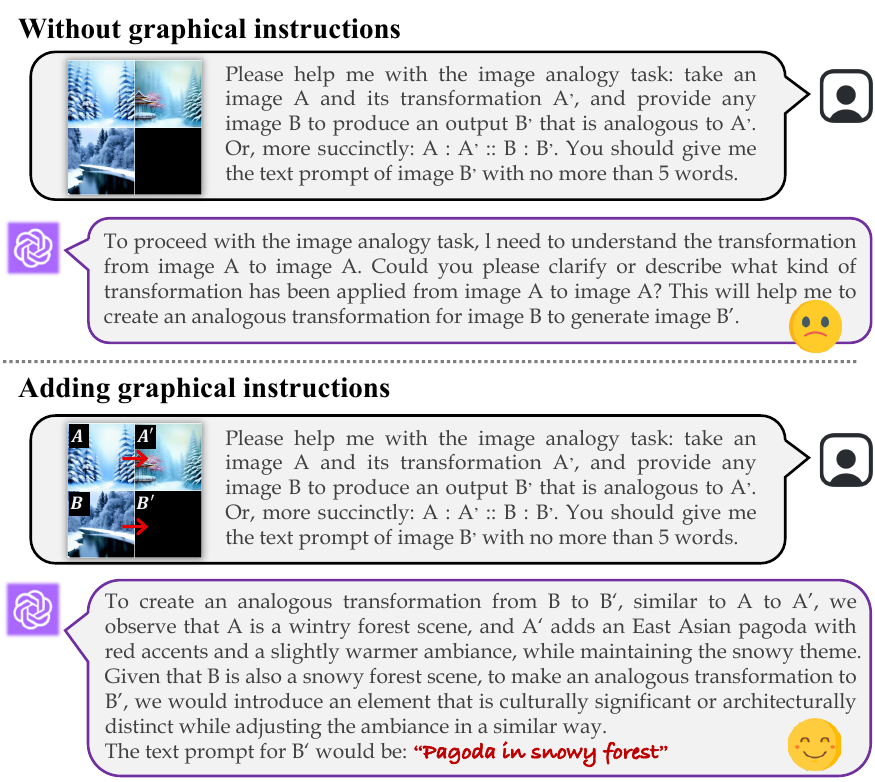}
    \caption{Ablation on the graphical instructions in GPT-4V prompting. By adding marks and arrows, the identity and relation of the task becomes more obvious, making it easier for GPT-4V to produce proper text prompt. Source images: InstructPix2Pix~\cite{brooks2023instructpix2pix}.}
    \Description{}
    \label{fig:gpt_prompting}
\end{figure}

\begin{figure}[t]
    \centering
    \includegraphics[width=\linewidth]{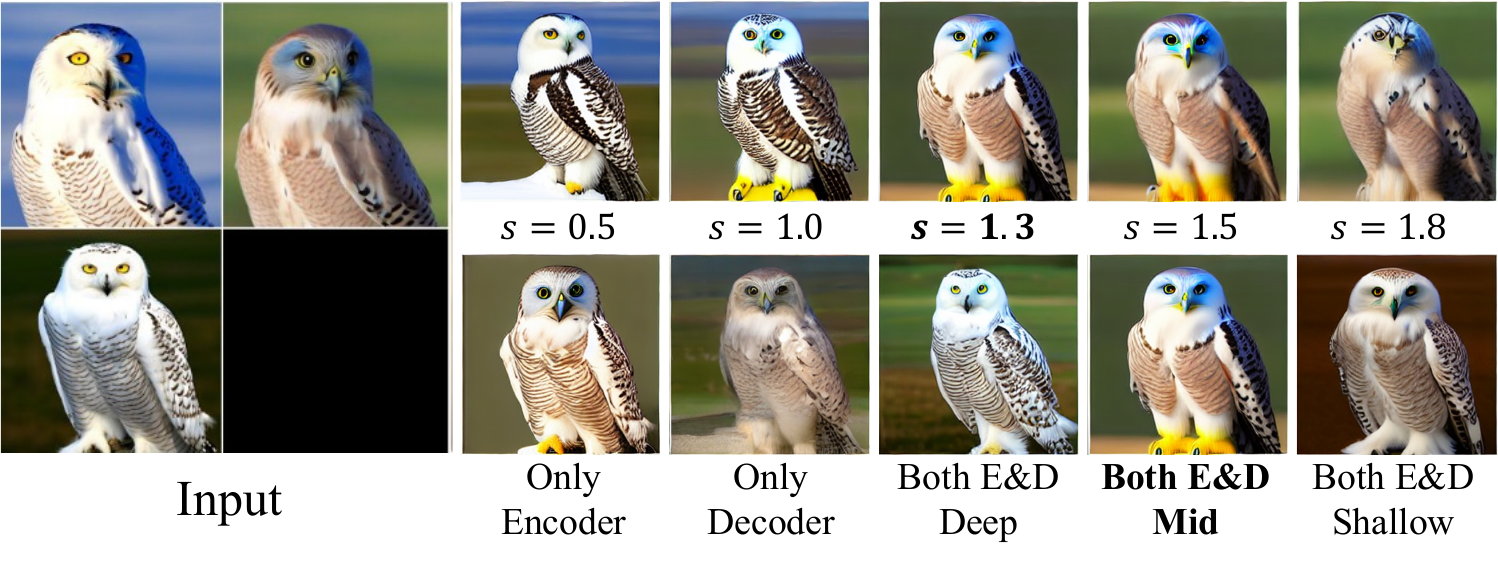}
    \caption{Ablation on hyper-parameters. In the first row, lower coefficient $s$ produces results more like $B$, while higher $s$ transfers more feature of $A'$. In the second row, performing SAC and CAM at middle layers ($16\times 16$) of the UNet achieves balance between structure preserving and transformation applying. Source images: InstructPix2Pix~\cite{brooks2023instructpix2pix}.}
    \Description{}
    \label{fig:hyper_paras}
\end{figure}

\subsection{Ablation Study}

\paragraph{Effectiveness of proposed components} To evaluate the effectiveness of the proposed components, we conduct a series of ablation studies. The ablation results are presented in Figure~\ref{fig:ablation}. (a) The baseline model of pretrained inpainting model generates rough and low-quality results. (b) {By pasting $B$ to the bottom right corner of the grid image}, the outputs are more structurally consistent with $B$. (c) Adding negative prompts helps to stabilize the generation process and avoid {messy} results. (d-1) Crucially, when operating self-attention cloning by $\mathcal{M}_s(B,B'):=\mathcal{M}_s(A,A')$, the model retains the information from $B$, but is unable to extract accurate context from $A'$ to infer the same transformation result. (d-2) When executing SAC by $\mathcal{M}_s(A',B'):=\mathcal{M}_s(A,B)$, the model is required to keep the structural relation between $A$ and $B$ consistent, after they have been transformed into $A'$ and $B'$. {Thus, we use (d-2) instead of (d-1).} (e) When adding textual prompts from GPT-4V {in} the whole grid image, the model rarely {focuses} the text guidance on the target inpainting area $B'$. (f) Finally, with the proposed CAM, our full approach not only maintained respectable generation quality but also successfully identified the necessary visual editing (adding sunglasses), effects (applying a cubist style), and transformations (changing church into mosque) for the ICL task.
\paragraph{GPT-4V Prompting} We ablate on the designed graphical instructions that used to hint GPT-4V in Figure~\ref{fig:gpt_prompting}. Without adding the visual marks on the grid image, GPT-4V may not know the corresponding relationship of the given images, therefore is unable to correctly analyze the content according to the instructions. By explicitly marking the positions of images ($A$, $A'$, $B$, and $B'$) on the constructed grid image, GPT-4V conveniently understands the information contained in the pictures. Meanwhile, the introduced arrows from $A$ to $A'$ and $B$ to $B'$ successfully demonstrate the transformation relations, making it more acceptable for GPT-4V to produce the ideal response of adding a ``pagoda in the snowy forest''. This text prompt will introduce semantic contextual information for the pretrained model to understand the task. {Note that our method is generic and supports other vision-language models~\cite{zhu2023minigpt} as well.}

\begin{figure}[t]
    \centering
    \includegraphics[width=\linewidth]{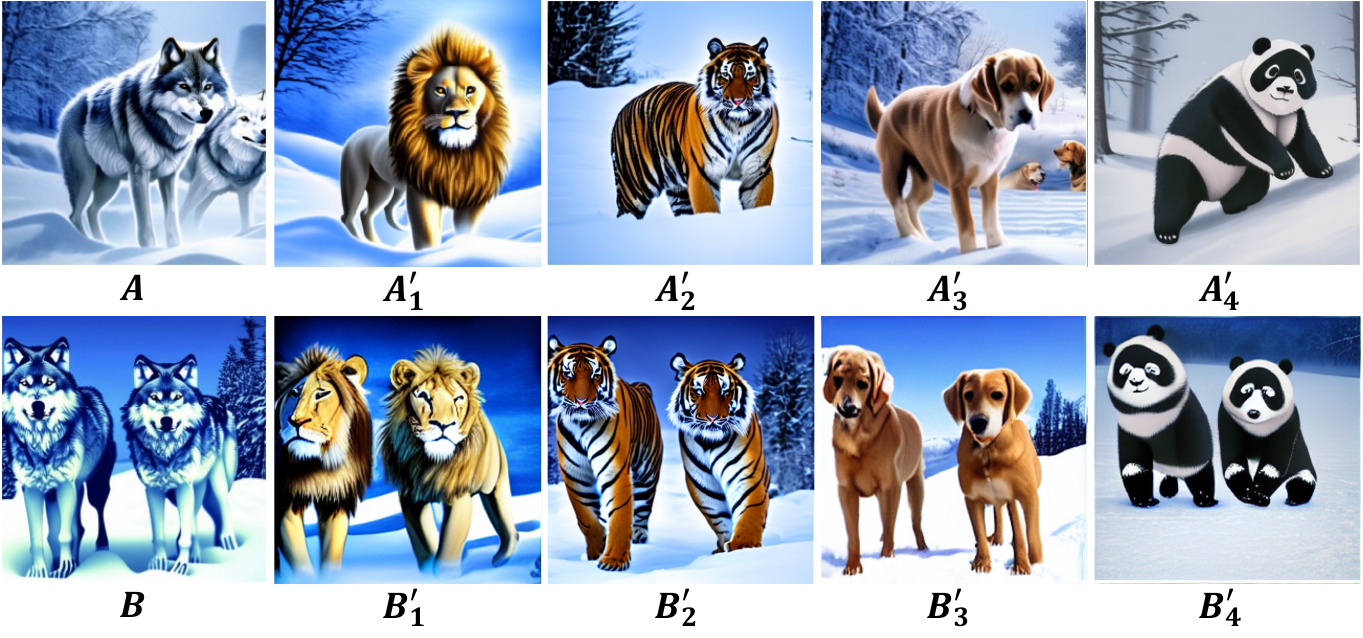}
    \caption{Given the same image $A$ and $B$ in the first column, and different $A'$s, our method is able to recognize the contextual relation between $A$ and $A'$ and produce the output $B'$ images accordingly. Source image: $A$ and $B$ are from ImageBrush~\cite{sun2023imagebrush}. $\{A_1',A_2',A_3',A_4'\}$ are generated using MasaCtrl~\cite{cao2023masactrl}.}
    \Description{}
    \label{fig:different_Ap}
\end{figure}

\begin{table}[t]
\centering
\caption{Comparison of inference time taken to perform one ICL task for different methods. Compared to existing methods, our method does not require training on a specific task and additional optimization.}
{\begin{tabular}{lc}
\toprule
\textbf{Method}          & \textbf{Inference time} \\
\midrule
MAEVQGAN~\cite{bar2022visual}        &    0.4s     \\
PromptDiffusion~\cite{wang2023incontext} &    4s     \\
DIA~\cite{vsubrtova2023diffusion}             &    258s     \\
VISII~\cite{nguyen2023visual}           &    685s \\
\midrule
Analogist (ours)        &    4s    \\
\bottomrule
\end{tabular}}
\label{tab:runtime}
\end{table}

\paragraph{Hyper-parameters} We present ablation on the parameter sensitivity of our proposed method in Figure~\ref{fig:hyper_paras}. As for the SAC coefficient $s$, utilizing a smaller $s$ value ($s=0.5$) results in an output more closely resembling the original Image $B$, whereas a larger value ($s=1.3$) tends to imbue the result with characteristics of $A'$. However, excessively large coefficients ($s=1.8$) leads to an overly unbalanced attention map, which in turn reduces the quality of generation.
We also ablate the selection of UNet layers in which we perform SAC and CAM. The results indicate that it is necessary to perform operations simultaneously in both the encoder and the decoder. Furthermore, if the operations are performed at a shallow level (high resolution), the outcome is merely a simple replication of some colors and coarse textures, leading to poor quality. If the operations are performed at a deeper level (low resolution), the excessive compression of information leads to the generated result being similar to the original image $B$. In our experiments, we perform SAC and CAM at a middle level of the UNet layers.

\subsection{Analysis}


\paragraph{Different In-context examples}
A model with contextual reasoning abilities should be able to produce different results based on different in-context examples, when given the same input. To verify that our approach has such capabilities, we conducted the following experiment as shown in Figure~\ref{fig:different_Ap}. Given the same image $A$ as an image of wolves, we first translate $A$ into different example outputs $\left\{ A'_1, A'_2, A'_3, A'_4 \right\}$ using MasaCtrl~\cite{cao2023masactrl}, obtaining different animals like lion, tiger, dog, and panda. We construct different ICL tasks, keeping the image $A$ and $B$ being the same, while varying the image $A'$s. Our method is able to recognize the translation from $A$ to $A'$ accordingly and generate the corresponding animals in $B'$, demonstrating the ICL capacity of our Analogist.


\paragraph{Inference Runtime}
In this section, we compare the execution time for different ICL methods performed once. Our experiment is conducted on an RTX 3090 GPU, and we calculated the time taken to generate one image. The result is shown in Tab~\ref{tab:runtime}. MAEVQGAN~\cite{bar2022visual} is the least time-consuming, taking 0.4 seconds, since it is generating very few tokens without the need of iteratively denoising. Our method Analogist takes about 4 second, the same as PromptDiffusion~\cite{wang2023incontext}, which is typically the standard sampling time for Diffusion models, but does not require specific fine-tuning. As for the previous inference-baesd methods DIA~\cite{vsubrtova2023diffusion} and VISII~\cite{nguyen2023visual}, it takes rather long time (\textit{i.e.}, 258 seconds and 685 seconds) for these two methods to estimate the CLIP feature and editing instruction respectively.

\begin{figure}[t]
    \centering
    \includegraphics[width=0.99\linewidth]{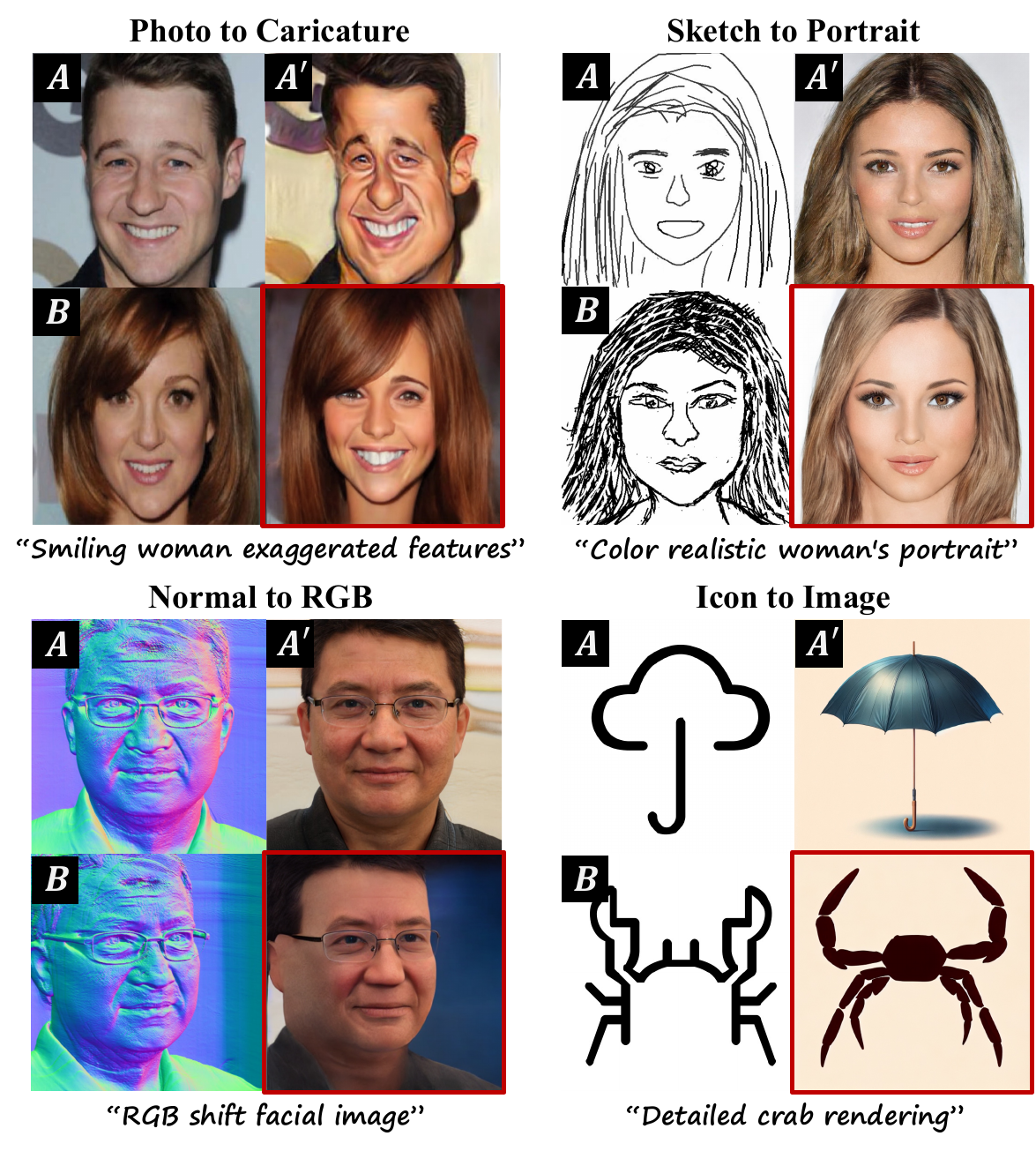}
    \caption{Examples of application for tasks where $A$ and $A'$ are aligned. The text prompts generated by GPT-4V is shown below each example. utput images are highlighted. Source image: Photo-to-caricature images are from CariMe~\cite{gu2021carime}. Sketch-to-portrait images are from DeepFaceDrawing~\cite{chen2020deepfacedrawing}. Normal-to-RGB images are from Trevithick et al.~\shortcite{trevithick2024you}. Icon images are from IconShop~\cite{wu2023iconshop}.}
    \Description{}
    \label{fig:application_aligned}
\end{figure}

\begin{figure}[t]
    \centering
    \includegraphics[width=0.99\linewidth]{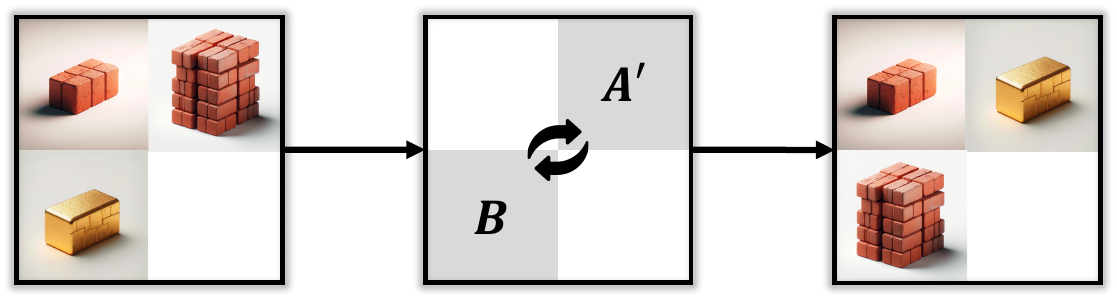}
    \caption{Illustration of the pipeline for tasks in which $A$ is aligned with $B$ instead of $A'$. We swap the positions of $A'$ and $B$ in the grid image. Through this way, we simplify the problem into aligned tasks. Source images: generated by DALLE-3~\cite{betker2023improving}.}
    \Description{}
    \label{fig:application_unaligned_pipeline}
\end{figure}

\begin{figure}[t]
    \centering
    \includegraphics[width=0.99\linewidth]{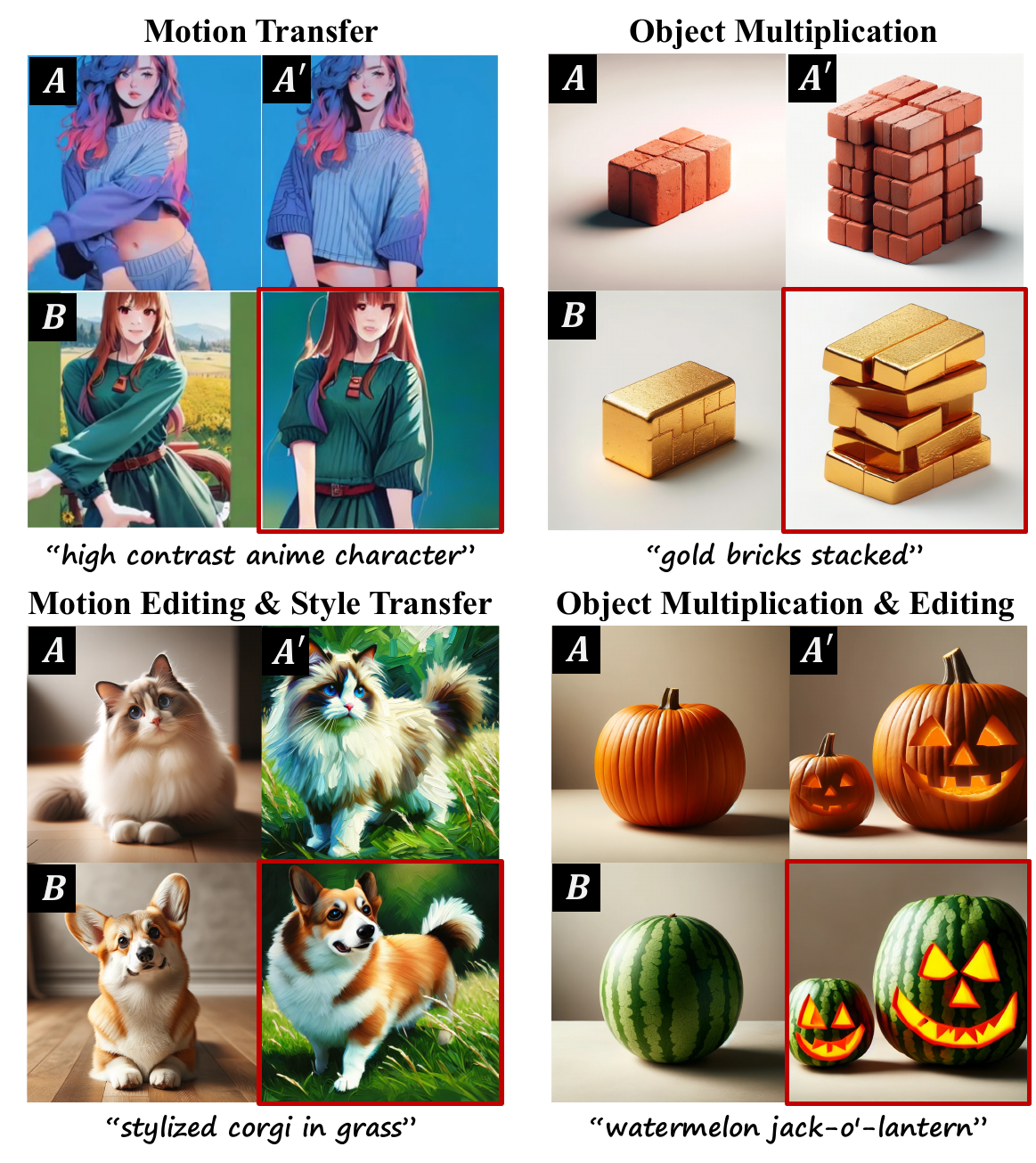}
    \caption{Examples of application for tasks where $A$ and $B$ are aligned. The text prompts of GPT-4V are shown below each example. Output images are highlighted. Source images: The example images of the first motion transfer case are from Chang et al.~\shortcite{chang2023magicdance}. The other three example images are generated by DALLE-3~\cite{betker2023improving}.}
    \Description{}
    \label{fig:application_unaligned}
\end{figure}

\begin{figure}[t]
    \centering
    \includegraphics[width=0.99\linewidth]{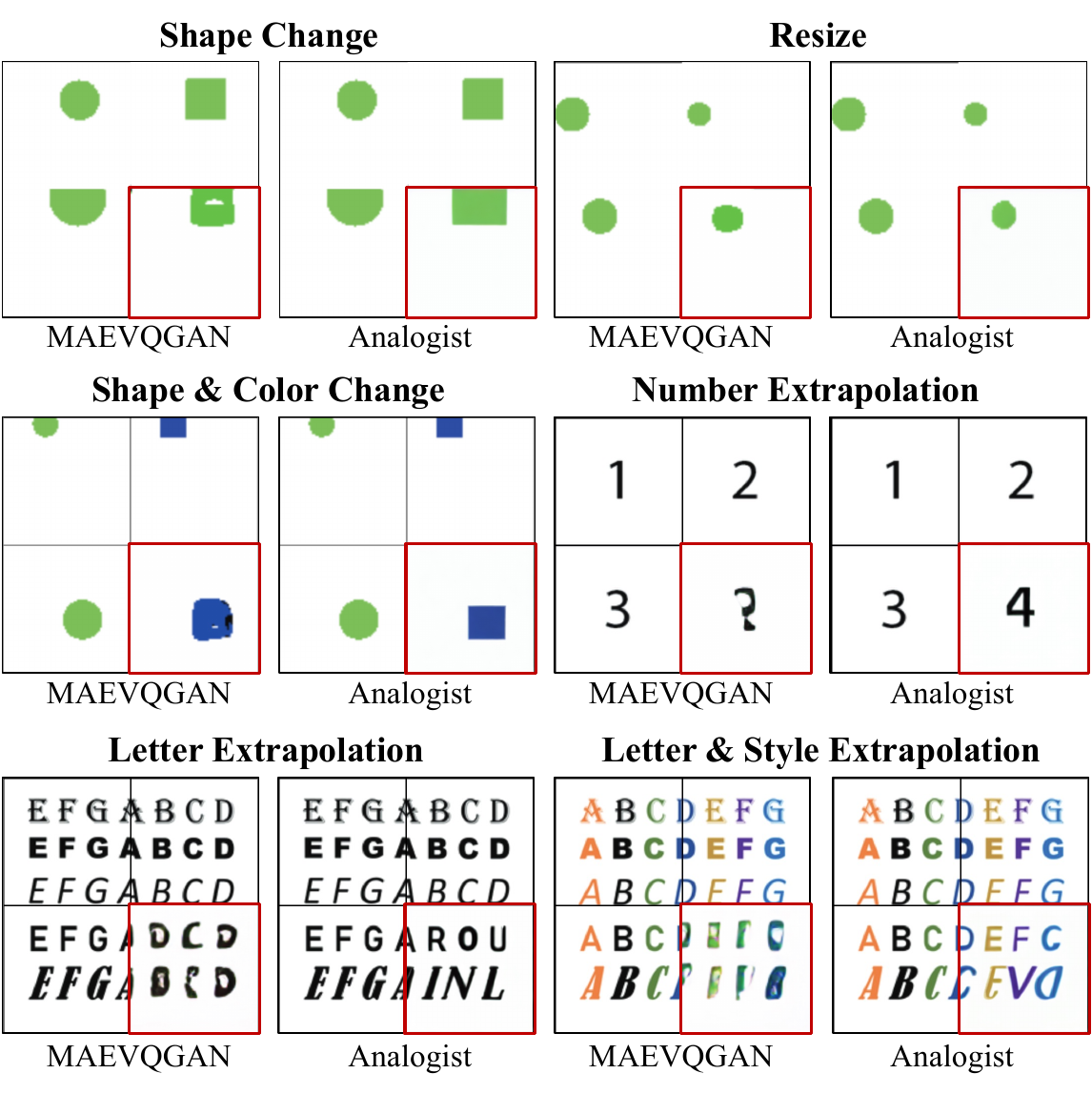}
    \caption{{Examples of application for tasks where $A$, $A'$ and $B$ are all misaligned. We test our method without SAC, only CAM is applied. Output images are highlighted. Source images: MAEVQGAN~\cite{bar2022visual}.}}
    \Description{}
    \label{fig:more_tasks}
\end{figure}

\section{Application}
{In this section, we extend Analogist to three categories of applications: (a) $A$ and $A'$ are aligned, (b) $A$ and $B$ are aligned, and (c) $A$, $A'$, and $B$ are all misaligned. For (b) and (c), we make adjustments to our method accordingly.}

\subsection{\texorpdfstring{$A$}{Lg} and \texorpdfstring{$A'$}{Lg} are aligned}
Under the condition that $A$ and $A'$ are aligned, we show example of applications in Figure~\ref{fig:application_aligned}, \textit{e.g.}, photo-to-caricature, sketch-to-portrait, normal-to-RGB, and icon-to-image tasks. The results show that our method is able to generate reasonable results on these tasks. Notice that there are slight structural changes between $A$ and $A'$ for photo-to-caricature and icon-to-image. However, our method is still robust to these minor issues since we are providing in-context information from both structural and semantic levels.

\subsection{\texorpdfstring{$A$}{Lg} and \texorpdfstring{$B$}{Lg} are aligned}
We make it possible to address tasks where $A$ is aligned with $B$ instead of $A'$. We give an example of object multiplication in Figure~\ref{fig:application_unaligned_pipeline}, where $A$ contains one brick and $A'$ contains a brick stack. This problem can not be done through our original pipeline. To tackle this problem, we swap the positions of $A'$ and $B$ in the grid image, constructing a new grid image where $A'$ contains one brick and $B$ contains a stack of bricks. In this way, we simplify the task into one where $A$ and $A'$ are aligned again, \textit{i.e.}, changing the task of \textit{turning one brick into brick stack} into the task of \textit{changing bricks into golden bricks}. This strategy can be applied to tasks like motion transfer and image analogy where $A$ and $A'$ are misaligned in figure~\ref{fig:application_unaligned}. We also demonstrate our method's ability of addressing tasks with multiple translations like both motion editing and style transfer, and object multiplication with editing.

\subsection{\texorpdfstring{$A$}{Lg}, \texorpdfstring{$A'$}{Lg}, and \texorpdfstring{$B$}{Lg} are all misaligned}
\label{sec:more_tasks}
{We extend our method on tasks where $A$, $A'$, and $B$ are all misaligned in Figure~\ref{fig:more_tasks}, such as changing a circle to a square, resizing a big circle to a smaller one, extrapolating new content of numbers and letters. We test our method without SAC to prevent incorrect structure guidance. Analogist produces reasonable results and outperforms MAEVQGAN. It should be pointed out that the quality of long sequence letter generation still have room to improve due to notorious tendency of diffusion models to struggle with generating high-quality text. Nevertheless, we believe these results demonstrate the pre-trained generative models have ample potential of in-context ability to be further tapped.}

\begin{figure}
  \centering
  \subfigure[Example of inaccurate prompt by GPT-4V. The expected right prompt is shown above the image with the critical words marked green. The prompt given by GPT-4V is shown below with the wrong words in red.]{
    \includegraphics[width=0.485\textwidth]{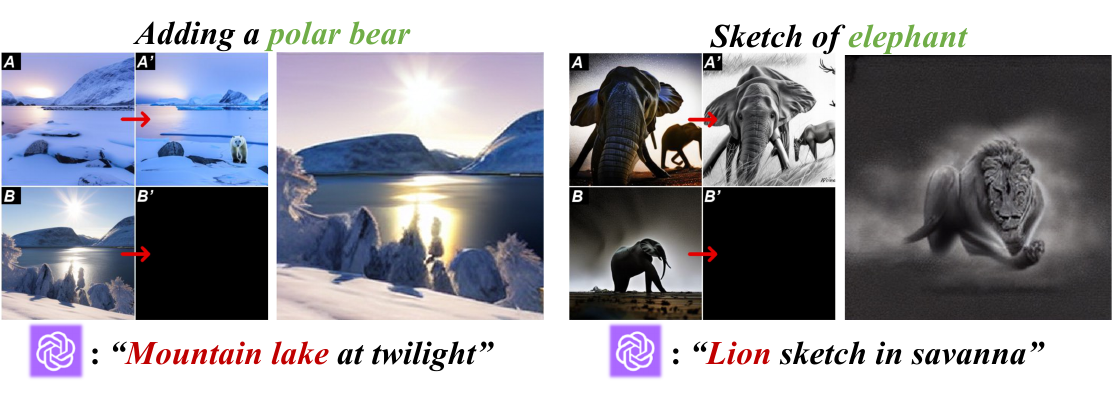}
    \label{fig:failure_1}
  }
  \subfigure[Failure examples of generating unnatural images on which the model is rarely seen during the pretraining stage, for example, normal maps and abstract icons.]{
    \includegraphics[width=0.485\textwidth]{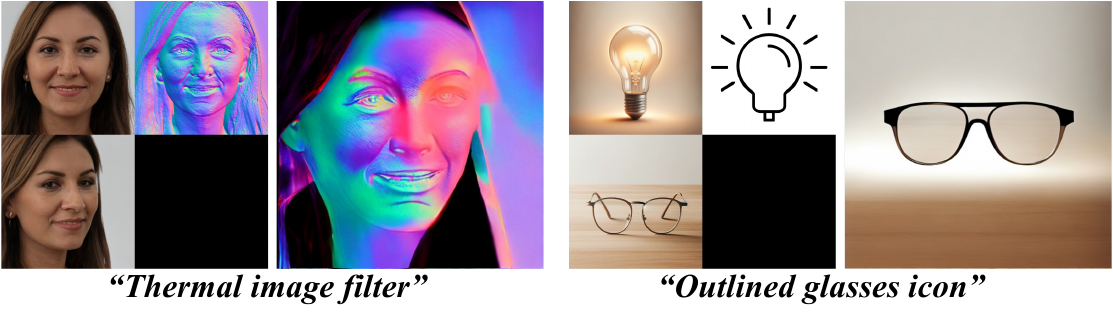}
    \label{fig:failure_2}
  }
  \subfigure[Example of $A$, $A'$, and $B$ are all misaligned, where SAC is not applicable.]{
    \includegraphics[width=0.485\textwidth]{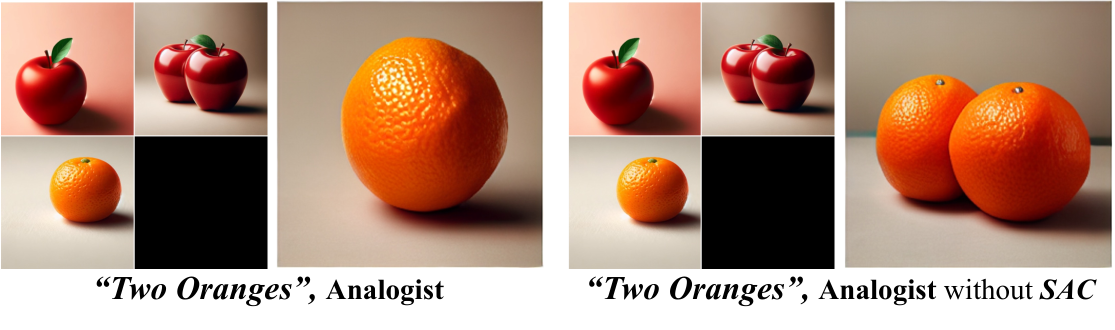}
    \label{fig:failure_3}
  }
  \caption{Example of failure cases. (a) GPT-4V fails to accurately deduce the correct textual prompt from the given grid images when the transformation (adding a polar bear) or category (elephant, instead of lion) is ambiguous. (b) The model fails to generate unnatural images like normal maps or icons even though given the right text prompt. (c) The proposed SAC struggles with tasks where $A$, $A'$, and $B$ are all misaligned. Source image: Trevithick et al.~\shortcite{trevithick2024you}, IconShop~\cite{wu2023iconshop}, and DALLE-3~\cite{betker2023improving}.}
  \Description{}
  \label{fig:failure}
\end{figure}

\section{Limitation}
Although our approach enhances in-context learning abilities, it's important to consider two possible limitations. Firstly, the inpainting model might be misled by incorrect text descriptions. In Figure~\ref{fig:failure_1}, when the transformation from $A$ to $A'$ is minor (\textit{i.e.}, the added object in the first case is small and easily overlooked), GPT-4V fails to recognize it. The second case shows an style transfer task of drawing ``a sketch of elephant''. However, GPT-4V recognizes the object as a lion instead of an elephant, leading to inaccurate guidance. The potential solution could be leaving an interface for users to monitor and customize the text prompts in real time.

Secondly, the model struggles with producing data that it seldom sees during the training stage. As shown in Figure~\ref{fig:failure_2}, when asked to produce unnatural images like normal map and line-drawing icons, the model fails to generate accurate results since most of its training data are natural RGB images. On the other hand, it explains our method's mediocre performance on vision tasks compared to ImageBrush~\cite{sun2023imagebrush}. We believe this could potentially be achieved by demanding {a more powerful pretrained base model}.

Finally, the proposed self-attention cloning may struggle with scenario in which $A$, $A'$, and $B$ are all misaligned as shown in Figure~\ref{fig:failure_3}. The structural-level information is not applicable in this case. One possible solution is to rely on semantic-level information to produce the transformation {as discussed in Section~\ref{sec:more_tasks}}.

\section{Conclusion}
Addressing the limitations of inaccurate instruction and tedious optimization of existing inference-based methods, we introduced Analogist, a novel approach for visual In-Context Learning (ICL) combining visual and textual prompting. The proposed method utilizes a text-to-image diffusion model pretrained for image inpainting, making it an out-of-the-box solution for a wide range of visual tasks. We innovate with Self-Attention Cloning (SAC) for visual prompting, enabling fine-grained structural-level analogy, and leverage GPT-4V's visual reasoning for efficient textual prompting, supplemented by Cross-Attention Masking (CAM) for enhanced semantic-level analogy accuracy. Our approach, without the need for extra training or optimization, demonstrates superior performance in both qualitative and quantitative measures, showcasing robust ICL capabilities.

\begin{acks}
This work was supported in part by the National Natural Science Foundation of China under Grant 62276128, Grant 62192783 in part by the Collaborative Innovation Center of Novel Software Technology and Industrialization, and a GRF grant from the Research Grants Council (RGC) of the Hong Kong Special Administrative Region, China [Project No. CityU 11216122].
\end{acks}

\bibliographystyle{ACM-Reference-Format}
\bibliography{bibtex}

\end{document}